\documentclass[times,twocolumn,final,authoryear]{elsarticle}

\usepackage{prletters}
\usepackage{framed,multirow}

\usepackage{amssymb}
\usepackage{latexsym}

\usepackage{url}
\usepackage{xcolor}
\usepackage{amsmath}
\definecolor{newcolor}{rgb}{.8,.349,.1}

\journal{Pattern Recognition Letters}

\begin{document}

\thispagestyle{empty}

\clearpage

\ifpreprint
  \setcounter{page}{1}
\else
  \setcounter{page}{1}
\fi

\begin{frontmatter}

\title{Saliency guided deep network for weakly-supervised image segmentation }

\author[1]{Fengdong \snm{Sun}} 
\author[1]{Wenhui \snm{Li}\corref{cor1}}
\cortext[cor1]{Corresponding author}
\ead{liwh@jlu.edu.cn}

\address[1]{College of Computer Science and Technology, Jilin University, Changchun, 130012, China}

\received{1 May 2013}
\finalform{10 May 2013}
\accepted{13 May 2013}
\availableonline{15 May 2013}
\communicated{S. Sarkar}

\begin{abstract}
Weakly-supervised image segmentation is an important task in computer vision. A key problem is how to obtain high quality objects location from image-level category. Classification activation mapping is a common method which can be used to generate high-precise object location cues. However these location cues are generally very sparse and small such that they can not provide effective information for image segmentation. In this paper, we propose a saliency guided image segmentation network to resolve this problem. We employ a self-attention saliency method to generate subtle saliency maps, and render the location cues grow as seeds by seeded region growing method to expand pixel-level labels extent. In the process of seeds growing, we use the saliency values to weight the similarity between pixels to control the growing. Therefore saliency information could help generate discriminative object regions, and the effects of wrong salient pixels can be suppressed efficiently. Experimental results on a common segmentation dataset PASCAL VOC2012 demonstrate the effectiveness of our method.
\end{abstract}


\end{frontmatter}

\section{Introduction}
Recently, computer vision research has a prominent progress and achieves excellent performance. Many tasks in computer vision field need plenty pixel-level annotations to guarantee the accuracy of the corresponding solutions, such as scene understanding \citep{Wang2018retrieve} and instance segmentations \citep{Wu2018deep}. The pixel-level annotations indicate that each pixel in the ground truth has a label referring to its category. However, it is very difficult to obtain such pixel-level annotation datasets, because this kind of annotations is time consuming and requires substantial ﬁnancial investments. The process of labeling a pixel-level ground truth generally consumes a subject several minutes. On the contrast, weakly-labeled visual data, which only indicate the categories included by images but do not provide the locations of these categories, can be obtained in a relatively fast and cheap manner. Therefore, it is important and meaningful to generate pixel-level annotation data using weakly-labeled images, i.e. weakly-supervised semantic segmentation\citep{Wang2018clusteringtip}. 

In this paper, we focus on conducting pixel-labeled segmentation using weakly-labeled data. However, there is a large performance gap between weakly and fully supervised image semantic segmentation \citep{Wu2018wherepr, Wu2018wheremultimedia}. A key problem is how to infer the objects locations according to image-level categories. \citep{Qi2016objectness} used objectness proposal information to guide a object localization network to generate location cues, then aggregating these cues to help semantic segmentation. Although there are lots of helpful information contained in these aggregated location cues. Meanwhile many interference informations are mixed into them. These interferences are difficult to distinguish and eliminate under weakly-supervision such that they may effect the accuracy of object localization and image semantic segmentation. \citep{Alexander2016SEC} employed a classification network to retrieve objects location cues based on classification activation maps. These location cues, which consist of some discriminative regions, are very reliable and robust that could be used to improve the performance of segmentation tasks. Therefore, \citep{Alexander2016SEC} used these location cues to train a semantic segmentation network immediately. However, the discriminative regions in location cues are too small and sparse that they do not have enough ability to tune the entire network\citep{Wu2018retrieval}.

\begin{figure}[!t]
	\centering
	\includegraphics[width=\linewidth]{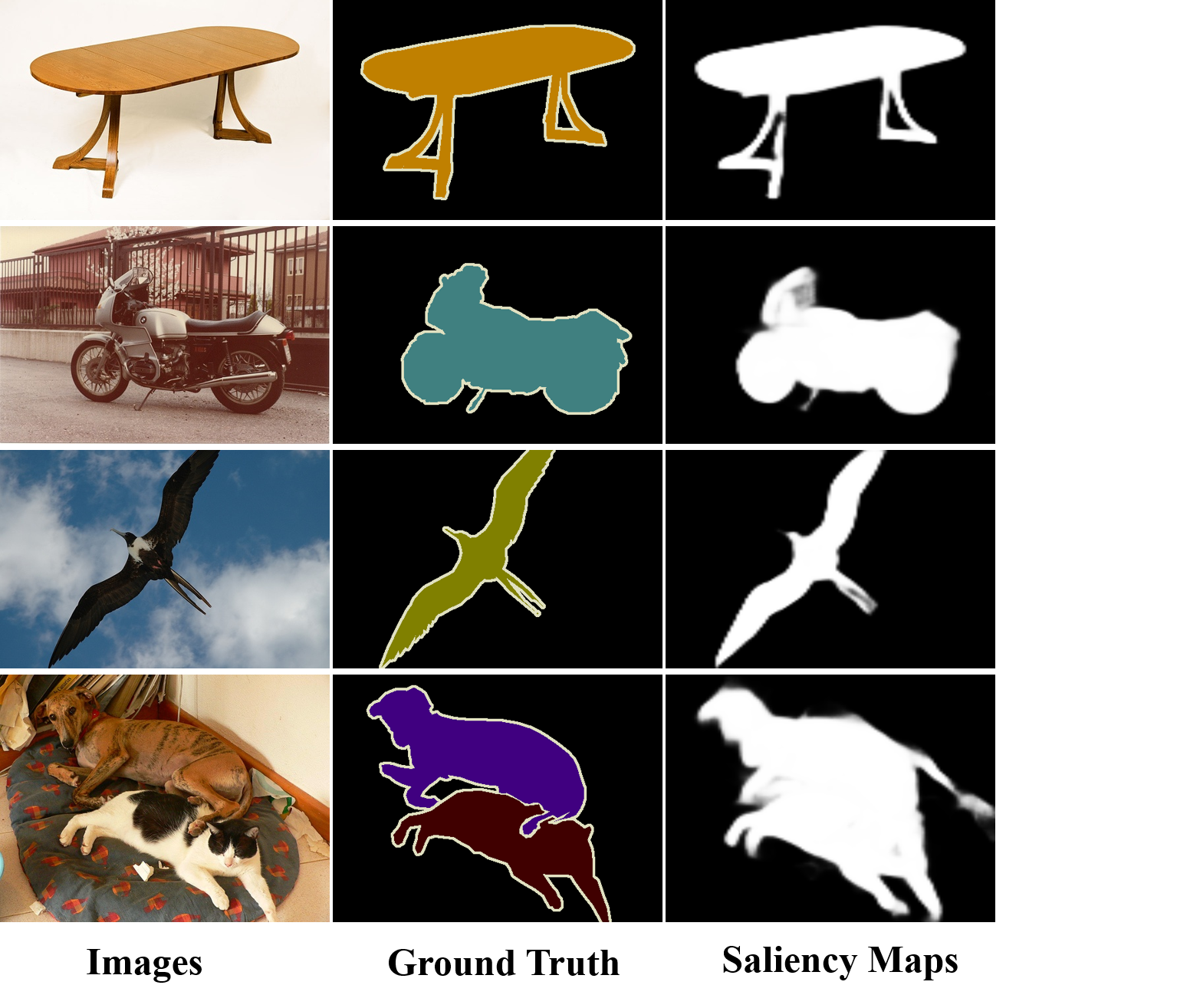}
	\caption{Some images in PACAL VOC 2012 validation set with their ground truths and saliency maps.}
	\label{fig:sal}
\end{figure}

For obtaining complete objects location from small and sparse cues, saliency detection methods are developed to enhance the performance of weakly-supervised segmentation. Saliency information of an image uses a saliency map to indicate the regions that most attract human beings' attentions. The saliency map can used to segment the input image into foreground and background. The salient foregrounds have a clear boundary and generally contain several salient objects, therefore could be utilized to generate objects locations from precise and reliable cues as Figure \ref{fig:sal} shown. \citep{Oh2017sal} propose to utilize saliency to assist semantic segmentation. However, they assign salient regions a category randomly picked from image-level labels initially, and these salient regions will be assigned to a category if the category's seed touching the salient regions afterwards. This method heavily depends on the precision of the saliency methods and may produce sub-optimal results if a salient regions just have an incorrect pixel touching a category seed.

To address the aforementioned problem, we propose a novel method called saliency guided weakly-supervised segmentation network which utilize saliency information as a guidance to generate robust objects locations from sparse cues to help image segmentation. Firstly, we use a weakly objection localization network to generate locations seeds from image-level category. These seeds have high confidence and precision that could be regard as ground truths. Secondly, to resolve the sparse and small issues of location seeds, we propose a novel method called saliency guided seeded region growing. The saliency information we used is from a self-attention saliency network which utilize image inherent cues, i.e. self-attention, to generate stage-wise refined saliency maps \citep{Sun2018selfattention}. We use the saliency detection method as a black box in this paper and immediately use the final saliency maps to guide the process of seeded region growing from location seeds. To alleviate the effect incorrect saliency results caused, we do not assign the salient region with a same label. We use the saliency values of pixels to generate saliency weights to control the process of seeded region growing. Therefore, a pixel not in a salient regions have possibility to get corresponding label, and a wrong salient pixel may be not grew by the location seeds. The saliency guidance can make the pixels with same saliency property easy to have the same label. At last, we integrate these cues into a network for weakly-supervised image segmentation. Experimental results demonstrate that our method outperforms several methods on a common PASCAL VOC2012 dataset.

In summary, the contributions of this paper are as followings: 
\begin{enumerate}
	\item We integrate weakly objects localization, saliency detection and saliency guided seeded region growing into a deep network framework for weakly-supervised segmentation.
	\item We propose a seeded region growing method with saliency guidance to expand the location generated by classification activation maps, therefore enriching pixel-level segmentation information.
	\item Experiments on a common dataset PASCAL VOC2012 demonstrate our method has a better performance than 11 existing algorithms.
\end{enumerate}

\section{Related Work}
\subsection{Saliency detection methods}
Many saliency detection methods are exploited for exact foreground segmentation recently\citep{Wang2018sctnnls}. In general, these methods can be divide into two categories: traditional methods and deep learning methods. Many researches demonstrate that deep learning methods have a significant improvement in accuracy of saliency detection, and various different modules are exploited to enhance the performance of deep saliency networks. \citep{Wang2017stagewise} propose a stagewise refinement model to refine the saliency maps. A coarse prediction map is generated by the model firstly. Then the a refinement structure is used to refine the coarse prediction map with local context information for a subtle saliency map. \citep{Zhang2018progressive} propose a progressive network which consists of multi-path recurrent connections and attention modules. The recurrent structure is used to improve the side-outputs of the backbone network. Then spatial and channel-wise attention mechanisms are used to assign more weights to foreground regions. \citep{Islam18revisiting} utilizes a framework to integrate three saliency tasks including detection, ranking and subitizing. They consider not only detect salient objects but also predict the total number and the rank order of them by the proposed framework. 
\begin{figure*}[!t]
	\centering
	\includegraphics[width=\textwidth]{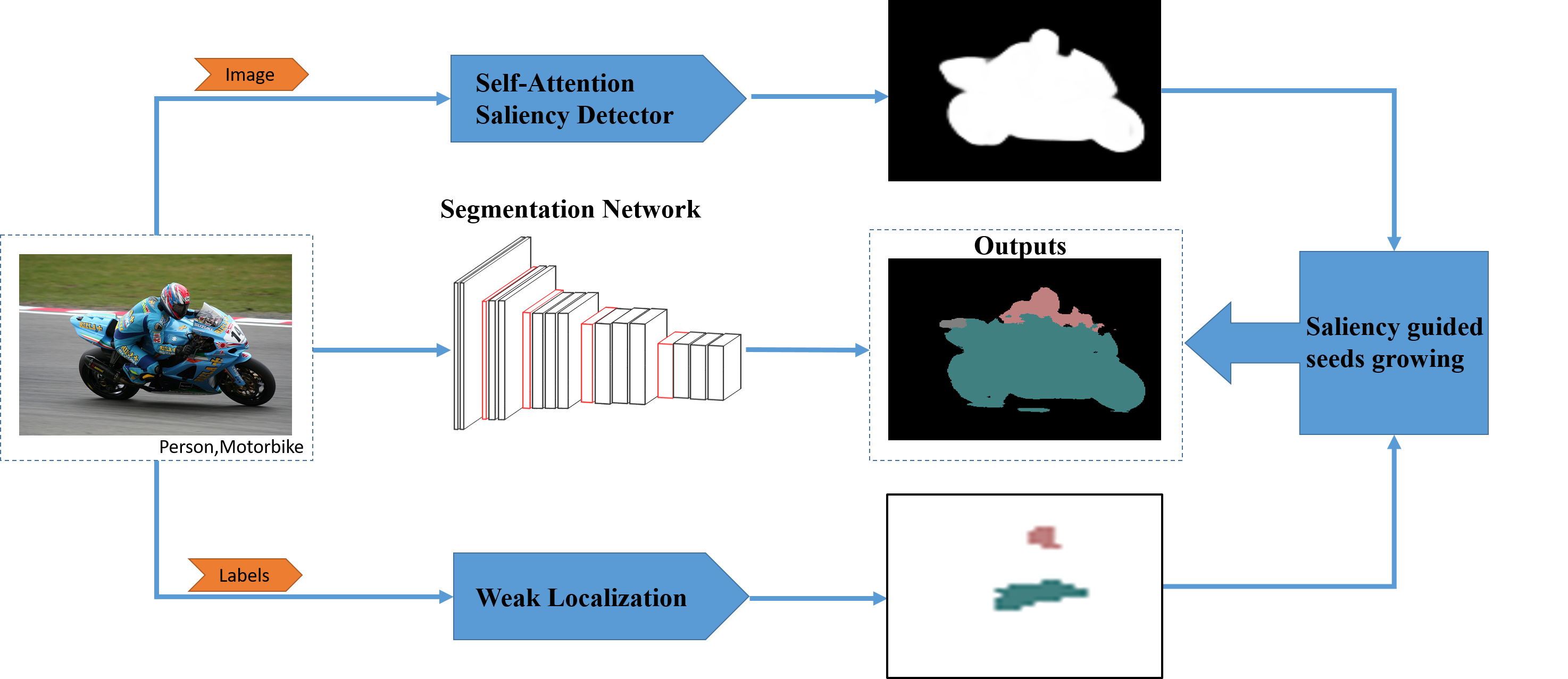}
	\caption{The overall framework of our network.The saliency guided seeds grown regions are used to train the segmentation network.}
	\label{fig:overall}
\end{figure*}

In this paper, we use a self-attention saliency network to conduct saliency detection. We utilize a self-attention module, which is calculated by the layer inputs, to enhance the salient semantics of deep layers from layer-level. A refined side-outputs by gated units are used to help network recover the resolution and generate exact saliency maps. The saliency maps can be segmented to the regions with clear boundaries, which can indicate objects locations. Therefore, we can use the saliency maps to enrich the location cues thus improving the performance of the weakly-supervised image segmentation.
\subsection{Weakly-supervised image segmentation}
Recently, many researches are emerging in weakly-supervised image segmentation \citep{Wu2018atten}. These researches achieve significant performance. There are some different kinds of weak labels, such as image labels\citep{Alexander2016SEC,Huang2018dsrg}, points~\citep{Amy2016points}, scribbles\citep{Lin2016scribble}. In this section, we mainly introduce the weakly-supervised segmentation from image labels. \citep{Wei2017erase} propose a  adversarial erasing approach to locate the object regions. The approach starts with a single small object region, then the region will be erased in an adversarial manner for discovering new and
complement object regions. An online learning
is also developed to enhance the adversarial erasing approach. \citep{Qi2016objectness} use objectness information to enhance the performance of weakly-supervised segmentation. One hand, The proposed method use a segmentation network to generate objectness proposals. On the other hand, the proposals are aggregated with object localizations to guide the segmentation network for a better performance. \citep{Oh2017sal,Chaud2017discovering} fuse saliency cues into weakly-supervised segmentation using different methods. \citep{Oh2017sal} use a existing saliency method to guide the process of training. \citep{Chaud2017discovering} exploit a novel saliency detection method, and combine saliency information with fully attention maps to segment input images. 

In this paper, we propose a novel saliency guided weakly-supervised network. Different with \citep{Oh2017sal}, we do not regard the pixels segmented by saliency method having the same class. First, classification activation maps are used to generate object location cues from image-level labels. These cues will be used as seeds which have high confidence and precision. Then the saliency information are utilized to help these seeds grow to enrich the object location regions. Therefore, we can use the grown regions to train the network for image segmentation.  

\section{Proposed Method}
In this section, we introduce the method proposed in details. First, we use a the classification activation maps of a weak object location network to obtain location cues from image-level label. Then the saliency cues guide the seeds, i.e. the location cues, to grow based on similarity of pixels such that obtaining more object locations. At last, a deep segmentation network is used to learn segment input image using grown object locations. 


\subsection{Overall structure}
\label{sec:overall}
In this section, we will illustrate the overall structure of our method. As shown in Figure \ref{fig:overall}, the entire network has three components. The first component is a self-attention network to generate saliency maps of inputs. The second component is a semantic segmentation network for segmenting the inputs into regions with different labels. The third component is a weak object localization network to generate location cues as seeds. Besides, a small module is used to conduct seeded regions growing under saliency guidance.

When images feed into the network with its categories, the images and category labels are handled by different component. The category labels are transferred to the weak object location network to generate sparse but reliable location cues. The network utilize classification activation maps, which fuse the last convolutional feature maps with their response weights to the images' categories, to extract the location cues. These cues have a high precision but many of them are scattered. To address this problem, we use a saliency guided region growing method to extend the location cues. A saliency detection method based on self-attention are utilized to produce saliency maps which are used to help extend location cues. The saliency network will assign each element in deep layers a self-attention weight to emphasize salient foreground pixel and alleviate the interference of background regions. Then a saliency guided seed region growing method can be utilized to extend location cues. The growing method not only considers the similarity between pixels but also takes account of their saliency values thus can obtain dense location labels.

Therefore, the results of segmentation network could be supervised by the dense location labels from the seeded region growing method. In the segmentation network, we use a modified VGG16 model pre-trained on ImageNet dataset. The last fully convolutional layer are used to conduct segmentation by a softmax function. For making the boundaries of segmentation more clear, we construct a fully-connected conditional random fields with unary potentials given by the predictions, and pairwise potentials of fixed parametric form which is from input images pixels. In this way, the segmentation network could classify each pixel's category of input images from image-level labels.    

\subsection{Seeds generation from classification activation maps}
\label{sec:seeds}
We utilize a deep network to detect discriminative object locations as seed cues under image-level labels. Recently, there are many different methods proposed to locate object regions from image-level label, such as multiple instance learning \citep{Pinheiro2015mil}. Driven by the progress of deep learning, many researches have focused on predict object locations with a deep network. And it can be observed that high-quality seeds, i.e. discriminative object regions, can be obtained by the feature maps of a classification network under the supervision of image-level categories. \citep{Zhou2016deepfeature} propose a fully convolutional classification network to predict seed regions using classification activation maps from image category. These activation maps from deep layers generally contain abundant object regions information for robust object localization. Therefore, we employ the seed generation method using classification activation maps.

The input images are feed into a network which is the modified VGG16 network. In this network, the fully connected layers of VGG16 are removed and we use conv7 to represent the last convolutional layers before the final output layer for convenience. The feature maps in conv7 contain abundant location information that are not utilized effectively. A global average pooling (GAP) are used to calculate the spatial average of each feature map in conv7. Then a weighted sum of these GAP values is used to generate the final output, i.e. the image-level category. These weights represent the importance of the GAP values of different feature maps to the image-level category. Therefore, they also can be used to weight the feature maps in conv7 thus helping identify the importance of different image regions to the image category. Then the regions with high importance are used as object location cues, and the weighted feature maps are the classification activation maps

For a given image, the $k^{th}$ feature map in conv7 can be represented as $f_{k}$, then $f_{k}(x,y)$ denotes the value at location $(x,y)$. The the result $F_{k}$ of the $k^{th}$ feature map after global average pooling  is as following:
\begin{equation}
F_{k} = \sum_{(x,y)} f_k(x,y)
\end{equation}
Thus, for a given category $c$, the value will be feed into softmax can be represented as:
\begin{equation}
S_c = \sum_{k}{w_k^cF_{k}}
\end{equation}
where $w_k^c$ is the weight of $F_{k}$ for the image category $c$, indicate the importance of the $k^{th}$ feature map for the image category $c$. Therefore, the classification activation maps $M_c$ for category $c$ is given by
\begin{equation}
M_c(x,y) = \sum_{k}w_k^{c}f_k(x,y)
\end{equation}
Thus, $M_c(x,y)$ directly indicates the importance of location $(x,y)$ leading to classify the input image to category $c$.

\subsection{Saliency guided seeded region growing}
\label{sec:grow}
The location cues generated from classification activation maps have a high precision and confidence. There exists a notable problem that these cues are very sparse and small. As reported in \cite{Huang2018dsrg}, only about 40\% pixels in the seeds have labels. Such sparse data can not have a significant improvement in semantic segmentation. Therefore, we want to extend these cues to obtain denser location information. A simple idea is grow the location cues as seeds to unlabeled regions, i.e. seeded region growing method. The seeded region growing method will choose some pixels as initial seeds which are generally selected following by low-level image property, such as color information and texture \citep{Wang2018scnn}. Then the method starts from a seed, and seeks the neighborhood to obtain homogeneous image regions by calculating the similarity between seeds and its neighbor pixels. Since the location cues have been generated by classification activation maps, we use these location cues as seeds to obtain denser regions.

Although the seeded region growing method could extend object locations effectively. It may produce error grown pixels because there is lack of constraint condition when seeds growing. For example, the object seeds may grow to a background pixel if the pixel is adjacent to the seeds and has a similar appearance with the seeds. This may cause over-segmentation of discriminative regions. What's more, if background pixels are labeled as object regions, they may grow to adjacent homogeneous regions, i.e. other background pixels. This may affect the quality of segmentation. Therefore, we propose a saliency guided seeded region growing method. Saliency information is an image inherent property and generally presented by saliency maps which indicate the saliency value of each pixel. Salient regions segmented from saliency maps have a clear boundary such that can be used to guide the process of seeds growing. 
\begin{figure}[!t]
	\centering
	\includegraphics[width=\linewidth]{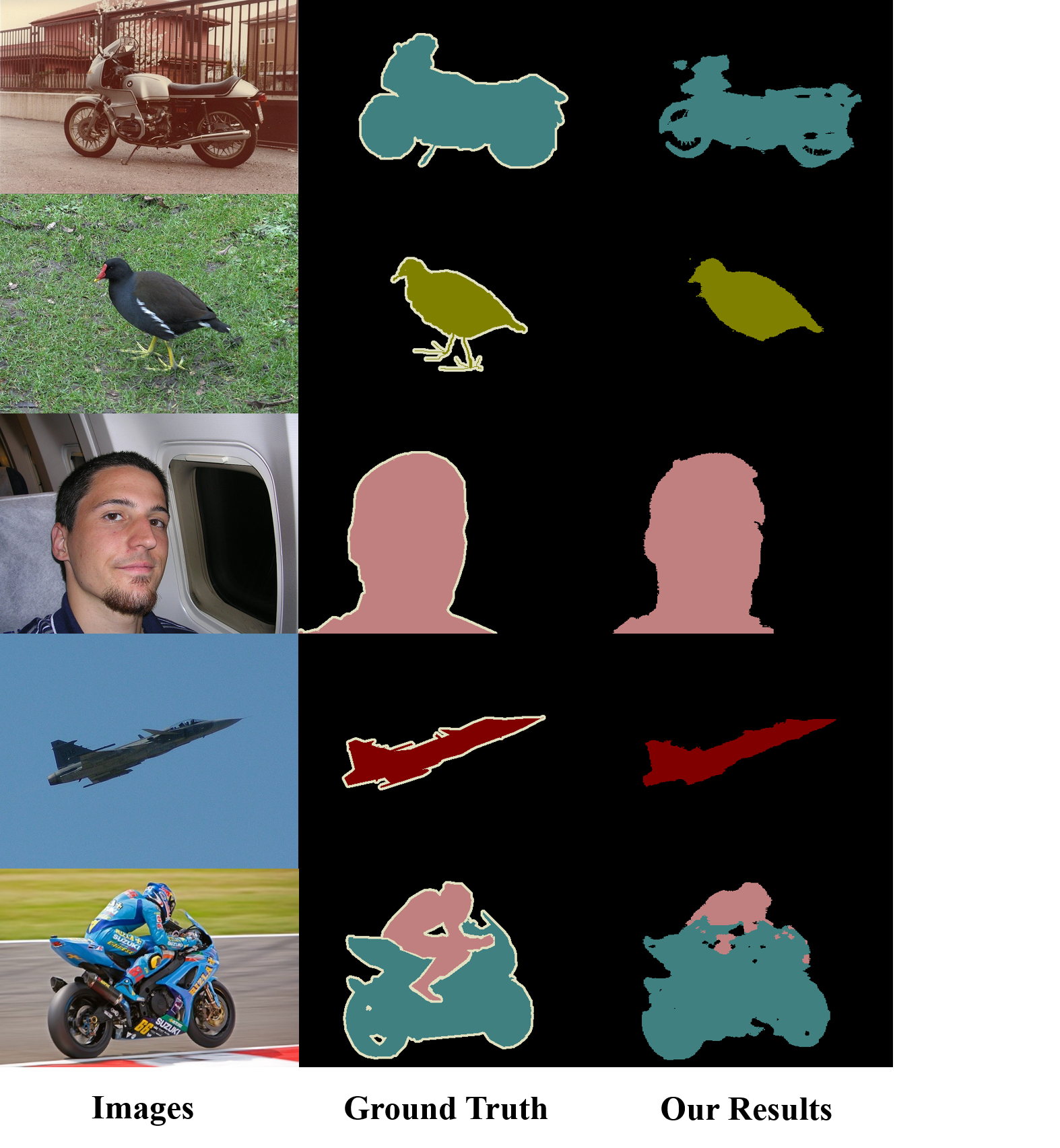}
	\caption{The results of our methods with ground truths.}
	\label{fig:result}
\end{figure}
\begin{table}
	\caption{\label{tab:1}Comparison of different methods on PASCAL VOC 2012 validation and testing sets (mIoU in \%).}
	\centering
	\renewcommand\arraystretch{2}
	\resizebox{8.5cm}{!}{%
		\begin{tabular}{lccc}
			\hline
			Method & Training Images & Val & Test\\
			\hline
			MIL-FCN \citep{MILFCN2014Pathak} & 10K & 25.7 &24.9\\
			\hline
			CCNN \citep{CCNN2015Pathak} & 700K & 35.3& 35.6\\
			\hline
			MIL-bb \citep{Pinheiro2015mil}&700K&37.8&37.0\\
			\hline
			EM-Adapt \citep{EMAdapt2015Papandreou}& 10K&38.2&39.6\\
			\hline
			SN-B \citep{SNB2016Wei} & 10K & 41.9 & 40.6\\
			\hline
			MIL-seg \citep{Pinheiro2015mil} & 700K & 42.0 & 43.2\\
			\hline
			DCSM \citep{DCSM2016Wataru}&10K&44.1&45.1\\
			\hline
			BFBP \citep{BFBP2016Fatemehsadat}&10K&46.6&48.0\\
			\hline
			STC \citep{STC2017Wei}&50K&49.8&51.2\\
			\hline
			Ours&10K&50.5&51.3\\
			\hline
			
		\end{tabular}%
	}
\end{table}

For a given image $I$ and its corresponding saliency map $S$. The similarity between two pixels $(x_i,y_i) and (x_j,y_j)$are defined as following
\begin{equation}
sim_{i,j} = w_{i,j}\left\| {I({x_i},{y_i}) - I({x_j},{y_j})} \right\| 
\end{equation}
where $I(x,y)$ represent the pixel value at location $(x,y)$, and $w_{i,j}$ is the saliency weights to control the groing. We use a HSV color space information to calculate the similarity. And the saliency weight $w_{i,j}$ are as following
\begin{equation}
w_{i,j} = exp(|S(x_i, y_i)-S(x_j, y_j)|)
\end{equation}
where $S(x,y)$ represents the value of the pixel at $(x,y)$ in the saliency map. Then the growing similarity criteria $P$ is given by
\begin{equation}
P_{i,c}\left( \theta \right) =
\begin{cases}
True\ \ \ \ \ \ \ \ \ if\ \ sim_{i,c}\ <\ \theta\\
False\ \ \ \ \ \ \ \ otherwise\\
\end{cases}
\end{equation}
where $c$ is the seed pixels with a label. If only the value of $P_{i,c}$ is true and pixel $i$ at location $(x_i,y_i)$ is adjacent to pixel $c$, the pixel $i$ can be assigned the same label with pixel $c$. The saliency weight will make the pixels with similar saliency can be broadcast easier such that the grown regions will accord with the shape of the salient objects.

\section{Experiments}
\subsection{Dataset and metrics}
We evaluate our model on the PASCAL VOC 2012 image dataset. There are several different tasks benchmark, and we use the segmentation class dataset to demonstrate the effectiveness of our method. The segmentation class dataset has three parts, including training set, validation set and testing set. The training set has 1464 images in total, and the other two sets have 1449 and 1456 images respectively. In a common practice, we augment the training set as the suggestion of Ref.~\citep{Har2011semantic}. Therefore, the final training dataset we used in this paper has 10,582 images with weak image-level labels. The validation and testing set are used to evaluate our method with other approaches. For the validation set, the ground truths are available such that we can use to generate the examples of prediction of our method. And for the testing set, the ground truths are not publicly available. Therefore, we submit the results of testing set to the official PASCAL VOC evaluation server to evaluate the performance of our method.

We adopt the standard intersection over union (IoU) criterion to evaluate a prediction and corresponding ground truth image \citep{Wang2018IJCAISC}. For a given image, $P$ and $G$ present the prediction image and ground truth image respectively. Then, the IoU of the prediction for this image can be given by
\begin{equation}
IoU = \frac{P\cap G}{P\cup G}
\end{equation}
We use the value of IoU to evaluate the performance in a image. Mean intersection over union (mIoU), which is the average IoU value of a dataset, can be used to evaluate the performance of a method in a dataset.
\begin{table*}[!t]
	\renewcommand\arraystretch{1.6}
	\caption{\label{tab:2}Detailed results of different method on PASCAL VOC 2012 dataset (mIoU in \%).}
	\centering
	\begin{tabular}{|c|c|c|c|c|c|c|c|c|c|c|} 
		\hline
		Val set& SFR &EM-adapt&CCNN&MIL-seg&Ours&Test set&CCNN&MIL-seg&RSP&Ours\\
		\hline
		background&71.7&	67.2 &68.5 &77.2&83.8&background&71&74.7& 74&84.7\\
		aeroplane&30.7&	29.2 &25.5& 37.3 &59.2&aeroplane&24.2 &38.8 &33.1 &58.5\\
		bicycle&30.5&	17.6 &18.0 &18.4&27.0&bicycle&19.9& 19.8& 21.7&27.0\\
		bird &26.3&28.6& 25.4 &25.4 &64.3&bird &26.3 &27.5&27.7 &66.2\\
		boat &20.0&22.2& 20.2 &28.2&26.4&boat &18.6& 21.7 &17.7&24.0\\
		bottle&24.2&29.6 &36.3& 31.9&39.0&bottle&38.1 &32.8 &38.4&45.7\\
		bus&39.2&47.0& 46.8& 41.6&67.4&bus&51.7 &40.0 &55.8&68.8\\
		car&33.7&44.0& 47.1& 48.1&57.9&car&42.9& 50.1 &38.3&54.3\\
		cat&50.2&44.2& 48.0& 50.7&71.8&cat&48.2& 47.1& 57.9&71.2\\
		chair&17.1&14.6& 15.8 &12.7&22.6&chair&15.6 &7.2& 13.6&22.7\\
		cow&29.7&35.1 &37.9& 45.7&52.5&cow&37.2& 44.8 &37.4&55.3\\
		diningtable&22.5&24.9 &21.0& 14.6& 24.4&diningtable&18.3 &15.8& 29.2& 22.6\\
		dog&41.3&41.0 &44.5& 50.9&62.6&dog&43.0& 49.4 &43.9&66.5\\
		horse&35.7&34.8& 34.5& 44.1&54.8&horse&38.2 &47.3 &39.1&59.0\\
		motorbike&43.0&41.6& 46.2& 39.2&60.8&motorbike&52.2 &36.6 &52.4&71.4\\
		person&36.0&32.1 &40.7 &37.9&53.8&person&40.0 &36.4& 44.4&55.3\\
		pottedpiant&29.0&24.8 &30.4& 28.3&35.0&pottedpiant&33.8 &24.3& 30.2&35.2\\
		sheep&34.9&37.4& 36.3& 44.0&63.6&sheep&36.0 &44.5& 48.7&58.7\\
		sofa&23.1&24.0& 22.2 &19.6&31.8&sofa&21.6 &21.0 &26.4&38.8\\
		train&33.2&38.1 &38.8 &37.6 &47.4&train&33.4 &31.5 &31.8 &39.9\\
		TVmonitor&33.2&31.6& 36.9 &35.0&51.8&TVmonitor&38.3& 41.3 &36.3&52.1\\
		\hline		
	\end{tabular}
\end{table*}
\subsection{Experiment settings}
The classification network we used to generate location cues is a slightly modified VGG16 network as the suggestions of Ref.~\citep{Alexander2016SEC}. The segmentation network we choose in this paper is the DeepLab-CRF-LargeFOV network which is introduced in Ref.~\citep{Chen2014segcrf}. The initial weights of these network are pre-trained on the ImageNet dataset~\citep{Deng2009imagenet}. Seeding losses introduced in Ref.~\citep{Alexander2016SEC} are used to calculate the losses between the segmentation outputs and the grown seeded regions. Stochasitc gradient descent optimizer is used for training the segmentation network with mini-batch. We use the momentum of 0.9 and a weight decay of 0.0005. The size of mini-batch is 4 and the weight decay parameter is 0.0005. We set a dropout rate 0.5 for the last two convolutional layers of segmentation network. The initial learning rate is 1e-3 and it is will be decreased by a factor of 10 every 10 epochs. 

In the seed generation, the pixels, whose values in the classification activation maps are in the top 20\%, are used as the object location cues. The corresponding saliency maps are generated by Ref.~\cite{Sun2018selfattention}. The parameter $\theta$ is the saliency guided seeded region growing method is set to 10. And we use the setting of Ref.~\cite{Philipp2011crf} to initialize the parameters of conditional random fields (CRFs). The CRFs are used to help generate final outputs of segmentation network, and recover the boundaries information of objects when upscaling the final output segmentations in the testing. 

\subsection{Comparisons with other methods}
We summarize some weakly-supervised image segmentation method, and show their results on PACAL VOC 2012 dataset in Table.~\ref{tab:1}, including MIL-FCN \citep{MILFCN2014Pathak}, CCNN \citep{CCNN2015Pathak}, MIL-bb \citep{Pinheiro2015mil}, EM-Adapt \citep{EMAdapt2015Papandreou}, SN-B \citep{SNB2016Wei}, MIL-seg  \citep{Pinheiro2015mil}, DCSM \citep{DCSM2016Wataru}, BFBP \citep{BFBP2016Fatemehsadat}, STC \citep{STC2017Wei}. The mIoU of different methods on validation set and testing set are shown with their training images. The table illustrates that our method has a highest mIoU score of these methods on both validation and testing datasets. We provide these results for reference and indicate the number of training images they used. Some methods are trained on different training sets or with different kinds of annotations, such as bounding boxes and image-level labels. Among the approaches, CCNN, MIL-bb and MIL-seg use a larger training set including 700K images. Mil-seg and SN-B implicitly utilize pixel-level supervision in the training phase.

Table~\ref{tab:2} shows the detailed results. The mIoU values of each category on validation and testing datasets demonstrate the effectiveness of our method. We compare our method with some methods including SFR \citep{SFR2016Kim}, RSP \citep{RSP2016Josip}, CCNN, MIL-seg. The mIoU values of our method are the highest in most categories. 
\subsection{Qualitative results}
Figure \ref{fig:result} shows some successful segmentation results. It
shows our method can produce accurate segmentations even
for complicated images and recover ﬁne details of the
boundary. It can be observed that the results of out method is very close to the ground truths. In the first four rows, we use four single objects to illustrate the effectiveness of saliency guidance. In the bottom row, there are two categories in the image, our method also can generate a satisfactory segmentation.
\section{Conclusion}
In this paper, we propose a novel method to segment images from image-level labels. The object location cues are generated by a localization network using the image category. Then a saliency guided seeded region growing method is used to extend these location cues. Therefore, the grown regions can be used to train a segmentation network for a better performance.

\section*{Acknowledgments}
This work was supported by the Science and Technology Development Plan of Jilin Province under Grant
20170204020GX, the National Science Foundation of China under Grant U1564211.
%
%
%
%
%
%
%
%

\bibliographystyle{model2-names}
\bibliography{article}

\end{document}